\documentclass{article}

    \usepackage[final,nonatbib]{neurips_2024}

\usepackage[utf8]{inputenc} 
\usepackage[T1]{fontenc}    
\usepackage{hyperref}       
\usepackage{url}            
\usepackage{booktabs}       
\usepackage{threeparttable}
\usepackage{amsfonts}       
\usepackage{nicefrac}       
\usepackage{microtype}      
\usepackage{xcolor}         
\usepackage{subcaption}
\usepackage{graphicx}
\usepackage{lipsum}
\usepackage{algorithm}
\usepackage{algorithmic}
\usepackage{amsmath}       

\title{Lateralization MLP: A Simple Brain-inspired Architecture for Diffusion}

\author{%
  Zizhao Hu\\
  University of Southern California \\
  \texttt{zizhaoh@usc.edu} \\  
   \And
   Mohammad Rostami \\
   University of Southern California \\
   \texttt{rostamim@usc.edu} \\
}

\begin{document}

\maketitle

\begin{abstract}

The Transformer architecture has dominated machine learning in a wide range of tasks. The specific characteristic of this architecture is an expensive scaled dot-product attention mechanism that models the inter-token interactions, which is known to be the reason behind its success. However, such a mechanism does not have a direct parallel to the human brain which brings the question if the scaled-dot product is necessary for intelligence with strong expressive power. Inspired by the lateralization of the human brain, we propose a new simple but effective architecture called the Lateralization MLP (L-MLP). Stacking L-MLP blocks can generate complex architectures. Each L-MLP block is based on a multi-layer perceptron (MLP) that permutes data dimensions, processes each dimension in parallel, merges them, and finally passes through a joint MLP. We discover that this specific design outperforms other MLP variants and performs comparably to a transformer-based architecture in the challenging diffusion task while being highly efficient. We conduct experiments using text-to-image generation tasks to demonstrate the effectiveness and efficiency of  L-MLP. Further, we look into the model behavior and discover a connection to the function of the human brain. Our code is publicly available: \url{https://github.com/zizhao-hu/L-MLP}
\end{abstract}


\section{Introduction}
Transformers~\cite{vaswani2017attention, dosovitskiy2020image} have become dominant in both natural language processing and computer vision. These architectures have demonstrated great scalability and flexibility and have replaced recurrent networks (RNNs) and convolutional networks (CNNs) in large-scale applications. The success of the Transformer is attributed to the incorporation of the multi-head self-attention mechanism. This mechanism uses dot-product attention to introduce inductive biases that are dependent on sequential interactions of the input tokens which enables dynamic parameterization. Despite the significant expressive power, the self-attention mechanism is computationally expensive and complicated.
Additionally, the self-attention mechanism does not have a similar counterpart in biological neural networks because the token-wise dot-product operations do not exist in the human brain. More specifically, the attention mechanism in the human brain is more likely done by a static nervous architecture rather than a dynamically parameterized one~\cite{raichle2015brain}. 
Since the human brain can inspire building human-level artificial intelligence systems,  this dissimilarity brings the question of whether the dynamic parameterization of self-attention is required for strong expressive power in AI similar to that of the human brain. In this work, we demonstrate that a lateralization-inspired architecture can achieve similar performance to the self-attention mechanism with parallels in the brain.


The left and right hemispheres of the human brain are generally symmetric in terms of basic structure, but they are functionally asymmetric which is referred to as the lateralization of the brain. For example, the left hemisphere is associated with language functions, while the right hemisphere is associated with more visuospatial functions. The two hemispheres each focus on specific aspects of external inputs and communicate with each other extensively through the corpus callosum.

We design a new architecture bearing the concept of symmetry in mind and mimicking the lateralization of the brain. Specifically, we investigate the performance of several existing MLP-based architectures in the context of diffusion tasks and find out that they fail to achieve comparable performance to Transformer-based backbones on a challenging text-to-image generation task. We discover that the design of L-MLP potentially closes this expressive gap between MLPs and Transformers and achieves comparable data generation quality. We demonstrate that a specifically designed U-shaped L-MLP (UL-MLP) can achieve comparable visual generations to a Transformer-based diffusion model at a lower cost.  Our backbone achieves an FID score of 8.62 on the MS-COCO validation with 20\% faster training and 40\% faster sampling speed. Our discovery also provides experimental evidence that functional lateralization can be formed in L-MLP during training, and the spatial-temporal processing lateralization is similar to that of a human brain. 

\section{Related Work}

The Vision Transformer (ViT)~\cite{dosovitskiy2020image} is increasingly taking over older models in computer vision. ViT splits images into patches and treating each patch as a token to form a sequence. Applying the self-attention mechanism to this sequence allows spatial information to be embedded through inter-token interactions and enables multimodal learning such as image-text and image-audio pair learning by concatenating multimodal tokens. Subsequent research has explored the trade-off between globality and scalability of ViTs~\cite{Liu2021SwinTH,Vaswani2021ScalingLS,Hassani2022NeighborhoodAT}. Other works focus on multimodal learning modification including cross-attention mechanisms~\cite{Jaegle2021PerceiverGP, Chen2021VisualGPTDA, Hu2021UniTMM}, adaptive layer normalization techniques~\cite{Xu2019UnderstandingAI}, and other hybrid architectures~\cite{Lu2019ViLBERTPT, Radford2021LearningTV}. These works bring various frameworks for processing both text and image data and set the basis for several Transformer-based diffusion models~\cite{Bao2022AllAW, Peebles2022ScalableDM, crowson2024scalable, Esser2024ScalingRF, Gu2021VectorQD, Yang2022YourVI}.

In recent years, MLP-based vision models~\cite{Tolstikhin2021MLPMixerAA, Liu2021PayAT, Touvron2021ResMLPFN, Yu2021S2MLPSM, Chen2021CycleMLPAM, Hou2021VisionPA} have emerged as a compelling alternative to traditional CNNs and ViTs. These models leverage the simplicity and efficiency of MLPs to achieve competitive performance in various vision tasks. MLP-Mixer~\cite{Tolstikhin2021MLPMixerAA} introduced the concept of token mixing and channel mixing using separate MLPs, demonstrating that convolutions and attention mechanisms are not strictly necessary for high performance in vision classification tasks. gMLP~\cite{Liu2021PayAT} incorporated gating mechanisms into MLP layers, enhancing gradient flow and expressiveness. resMLP~\cite{Touvron2021ResMLPFN} integrated residual connections into the MLP architecture, addressing the vanishing gradient problem and facilitating the training of deeper networks. S2MLP~\cite{Yu2021S2MLPSM} employed a spatial-shift operation to capture spatial relationships between pixels more effectively. CycleMLP~\cite{Chen2021CycleMLPAM} used cyclic shifting to capture long-range dependencies and contextual information more efficiently. Vision Permutator (ViP)~\cite{Hou2021VisionPA} proposed a permutation-based approach, allowing the model to permute input images and learn complex patterns.
These MLP-based models mostly focus on image classification tasks, where the requirement for the expressiveness of the model is not strong and moderate loss of information is negligible. However, for generation pixel-level tasks, such models have not been studied widely nor shown comparable performance to transformers. To  our knowledge, our proposed L-MLP is the first to demonstrate the generative capabilities of fully-MLP-based architectures on a challenging cross-modal text-to-image diffusion task. 

\section{Background and Preliminaries}

We introduce our MLP-based architecture in the context of text-to-image diffusion models.

\subsection{Diffusion models}


Score-based diffusion models are generative models that learn a to generate data from a probability distribution for a given dataset through modeling the process in which data samples diffuse from their latent representation. These models approximate the score function---a gradient of the log-probability of data---with a neural network. This approach avoids mode collapse and ensures diverse outputs compared to older generative models. The training objective is to minimize the mean square error between the true score function, \(S(\cdot)\), and its neural network approximation, \(s_\theta(\cdot)\), using:
\begin{equation}
L(\theta) = \mathbb{E}[(S(x_t) - s_\theta(x_t))^2], \quad \text{where} \quad S(x_t) = \nabla_{x_t} \log p(x_t),
\end{equation}
where $x_t$ is data sample, $S(\cdot$ is the scoring function, and $s_\theta(\cdot)$ denotes the neural network. 

For data generation, it is necessary to train a diffusion model and then draw samples from it.

\textbf{Training.}
The training process for a diffusion model involves two main steps: (i) Noise Addition, where data samples are progressively noised over multiple steps to create a fully noisy version of the input. (ii) Score function learning, where the network approximates the gradient of the log density of the noised data distribution to minimize the discrepancy over several iterations.

\textbf{Sampling.}
The sampling process reverses the noise addition to generate data samples from randomly drawn noise samples. To this end, the learned score function is used to gradually denoise and refine the data starting from the noisy initial state in the reverse steps used to train the model.

\textbf{Classifier-free guidance (CFG).}
Classifier-free guidance~\cite{Ho2022ClassifierFreeDG} models both the unconditional and conditional score functions to enable conditional generation. It does not require additional classifiers by incorporating the empty conditions for classifier-free training. It adjusts the score function as:
\begin{equation}
S(x_t|y) =  (1+\omega) \nabla_{x_t} \log p(x_t,y) - \omega \nabla_{x_t} \log p(x_t),
\end{equation}
where \( \omega \) is the guidance scale, facilitating a trade-off between fidelity and diversity.

\textbf{Latent Diffusion Models (LDMs).} 
LDMs~\cite{Rombach2021HighResolutionIS} operate in a latent space to efficiently generate high-definition images. They use a pretrained encoder to map images to a latent space and a decoder to reconstruct images from the denoised latent variables. This process significantly reduces the computational load compared to the classic diffusion model.

\textbf{Backbone architecture.}
Most existing diffusion models are Transformer-based or CNN-based. For guided diffusion models such as text-to-image diffusion models, cross-attention and adaptive layer-norm are common approaches. Transformer-based models suffer from low efficiency since the multi-step inferences in diffusion sampling exacerbate the computational complexity of such models. Thus, a simple and efficient MLP-based model can be a solution to these issues. Our proposed architecture relaxes the need for using attention layers by replacing them with MLP layers.

\section{Proposed Architecture}

Figure \ref{fig:pipeline} visualized our proposed architecture, called L-MLP, where a building block is presented in the left sub-figure.  L-MLP is a new MLP-based vision model, inspired by ViTs and ViPs. We adopt the same preprocessing step used by ViTs to split the images into patches. The L-MLP block's merging mechanism and residual connections are identical to those of a ViP model, except our model introduces separate normalization and global linear layers, which enables multimodal processing. For the last stage of processing, the L-MLP block uses an MLP identical to the one in a Transformer. We explain these components and the information processing pathways in the following sections.

\subsection{L-MLP}
\begin{figure}[h!]
  \centering
  \includegraphics[width=\textwidth]{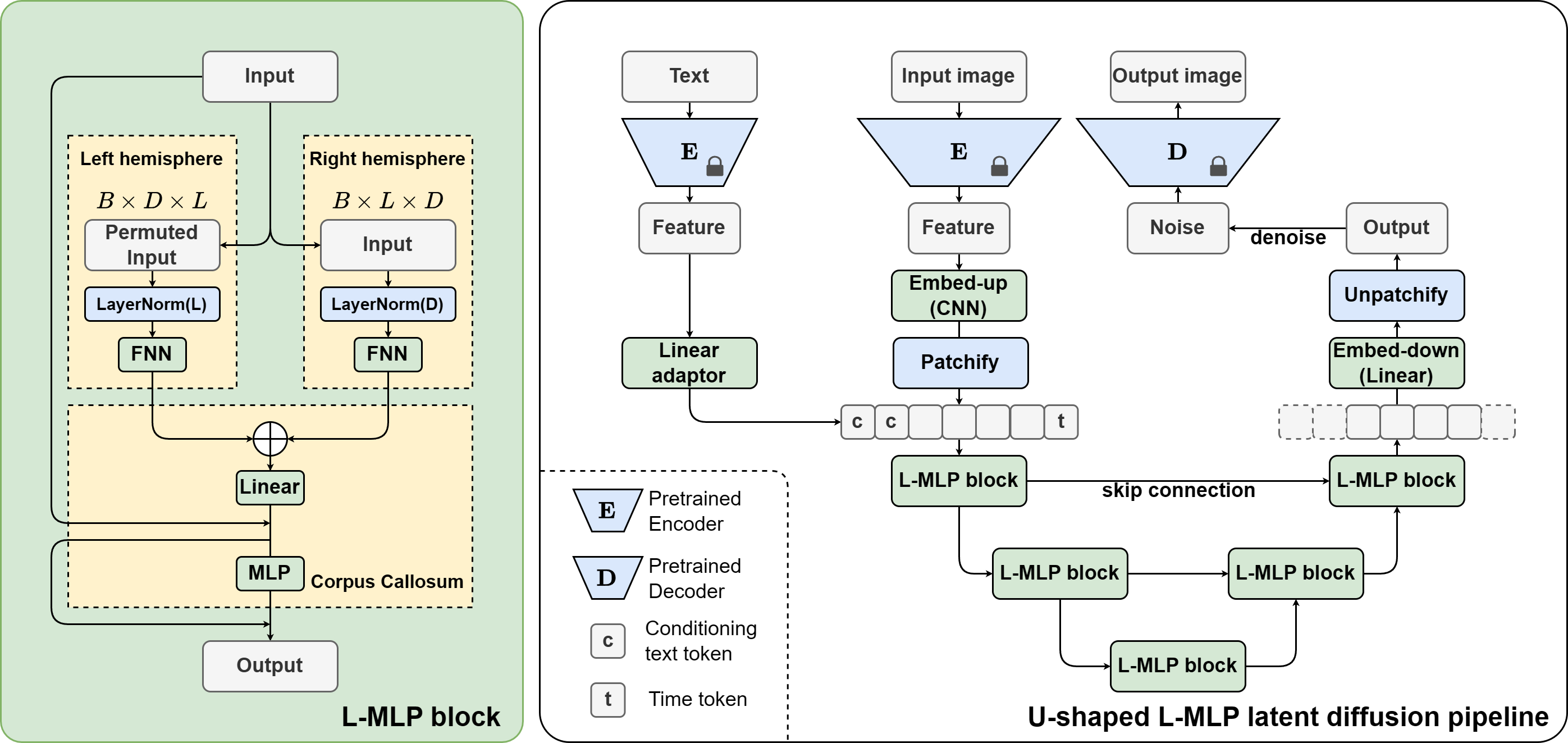}
  \caption{L-MLP block design and latent diffusion pipeline built from L-MLP blocks.}
  \label{fig:pipeline}
\end{figure}

We propose L-MLP, which is a simple block designed to enhance feature interaction across different dimensions and different modalities. The L-MLP block specifically processes input tensors of size $B \times L \times D$ similar to inputs of a Vision Transformer, where $B$ denotes the batch size, $L$ is the length of the sequence, and $D$ is the dimensionality of each sequence element. Our model introduces dimension permutation along the $L$ and $D$ dimensions, followed by separate normalization and transformations. Then the two pathways merge before being jointly processed along the $D$ dimension (see Figure \ref{fig:pipeline}, left, and Appendix A.2 for accurate implementation).

\textbf{Input permutation.} The architecture begins by permuting the last two dimensions of the input tensor $X \in \mathbb{R}^{B \times L \times D}$. The permutation transforms $X$ into a new tensor $X' \in \mathbb{R}^{B \times D \times L}$. Each of the tensors is then normalized along the last dimension.

\textbf{First stage processing.} Both the original input tensor $X$ and the permuted tensor $X'$ are separately fed into two feed-forward networks $\textit{FNN}_L$ and $\textit{FNN}_R$, where we use $L$ and $R$ to specify the left and the right feed-forward networks which are inspired by the two hemispheres of the human brain:
\begin{equation}
    R = \textit{FNN}_R(X)
\end{equation}
\begin{equation}
    L = \textit{FNN}_L(X')
\end{equation}
The processing of $X$ and $X'$ are governed by functional network $\textit{FNN}_R \in \mathbb{R}^{D \times D}$ and $\textit{FNN}_L \in \mathbb{R}^{L \times L}$, respectively. We use square mapping to maintain the shape of the input, facilitating the merging of branches in the next steps of data processing (see Figure~\ref{fig:pipeline}, right).

\textbf{Merging .} The merging is a simple addition followed by linear projection:
\begin{equation}
    Z = \textit{Linear}(L + R)
\end{equation}

\textbf{Second stage processing.} To merge the branches of the first stage processing, the processed inputs of the two feed-forward networks together with the residual connection from the input are then added before passing through another feed-forward network with an extra residual connection:  
\begin{equation}
    \textit{Output} = \textit{FNN}_C(Z + X) + Z + X
\end{equation}
This process leverages the interdependencies between the original and the permuted features through addition merging. The two-stage residual connections are arranged similarly to a transformer block. 

\textbf{Model Capacity.} To enable simple adjustment of the block capacity while maintaining the simplicity of the merging mechanism, we can change the size of the FFN hidden layer dimension through a scale factor. Here we have three potential adjustable FFN layers. Through our empirical explorations, we concluded adjusting the $\textit{FNN}_Z$ is sufficient. We also concluded that we can make $\textit{FNN}_R$ and $\textit{FNN}_L$ to be simple linear layers with a fixed scale factor equal to one to facilitate merging. 

\subsection{U-shaped L-MLP for latent diffusion}
Using L-MLP blocks, we design a U-shaped L-MLP (UL-MLP) network \ for solving the diffusion task (see Figure~\ref{fig:pipeline}, right). The general pipeline follows the latent diffusion model~\cite{Rombach2021HighResolutionIS}.

\textbf{Text encoder and adaptor.} We use a frozen CLIP~\cite{Radford2021LearningTV} text embedder to embed the captions. A linear adaptor is followed to scale up the embedding dimension size to 512.

\textbf{Image encoder, decoder, and adaptors.} We use a Stable Diffusion~\cite{Rombach2021HighResolutionIS} autoencoder to encode and decode the image. Followed by the encoder, there is a convolutional adaptor to scale up the embedding dimension size to 512. For the decoding process, we first use a linear adaptor to scale down the embedding dimensions back to 3, pass it through a convolutional layer, and then the Stable Diffusion decoder. In our empirical explorations, we noticed that the convolutional layers used in this process can be fully replaced with linear layers, resulting in a more strict MLP-only model. However, for a more fair comparison between the L-MLP block and the U-ViT transformer block, we stick with the setup used by U-ViT and use a convolutional network. These convolutional layers serve as preprocessing steps and are not included in the block design.

\textbf{Diffusion ODE solver.} We use DPM-Solver~\cite{lu2022dpm}
for the denoising process. For training, we train on $1000$ time steps. For sampling, we use $50$ steps for faster generation. 

\textbf{Long skip connections.}
Our backbone design follows the same concept as U-Net~\cite{Ronneberger2015UNetCN}, where the encoder UL-MLP blocks and decoder UL-MLP blocks are connected through skip connections to enhance the effect of early layers in data generation. We use simple addition for merging skipped inputs. We find that this design is contributing to better performance and faster learning.  
\section{Experiments}

Theoretically, our architecture can be easily adopted in many tasks, subject to relatively trivial changes.
However, we select the text-to-image diffusion task to study the model architecture for three advantages that classical classification tasks cannot provide: (i) it requires multimodal processing of text and image information, which provides a more challenging learning scenario, (ii) a generation task can facilitate the comparison between models through visual generations,  (iii) The multistep inference of diffusion models propagates errors that lead to diverse model behaviors. When combined, a small change in the model architecture will lead to drastically different visual generations from the same noise input which allows us to understand the connections between models. \textbf{Our implemented code is available: \url{https://github.com/zizhao-hu/L-MLP}.}

\subsection{Experimental settings}


We examine the performance and efficiency of the UL-MLP by applying them to a text-to-image generation task on the MS-COCO dataset~\cite{Lin2014MicrosoftCC}. The dataset contains $256\times 256$ images, with 82783 for training and 40504 for validation. Each image has 5 captions, and we select a random caption for training. For evaluation, we select $30000$ captions from the validation set to generate images. The image quality is then evaluated by the FID score~\cite{Heusel2017GANsTB}. The text-image alignment is evaluated by the CLIP score~\cite{Hessel2021CLIPScoreAR}. We adopt the diffusion model implemented by U-ViT, where timestamp embedding, conditional texts, and image feature inputs are all treated as tokens. We use the same preprocessing steps as U-ViT, where image features from a pretrained autoencoder are upscaled by a $2 \times 2$ patch convolutional layer to 512 dimensions. Text conditions are converted to an embedding of size $77 \times 512$ by a pretrained CLIP model followed by a linear adapter. We concatenate the caption embedding to image feature patches with a probability of $0.9$, and an empty text embedding otherwise, to enable classifier-free guidance training. The main model is trained on a single RTX 4090 for $7$ days with a batch size of $128$, and a gradient accumulation step of $2$.

\subsection{Ablative experiments for design choices}
\begin{figure}[h!]
  \centering
  \includegraphics[width=\textwidth]{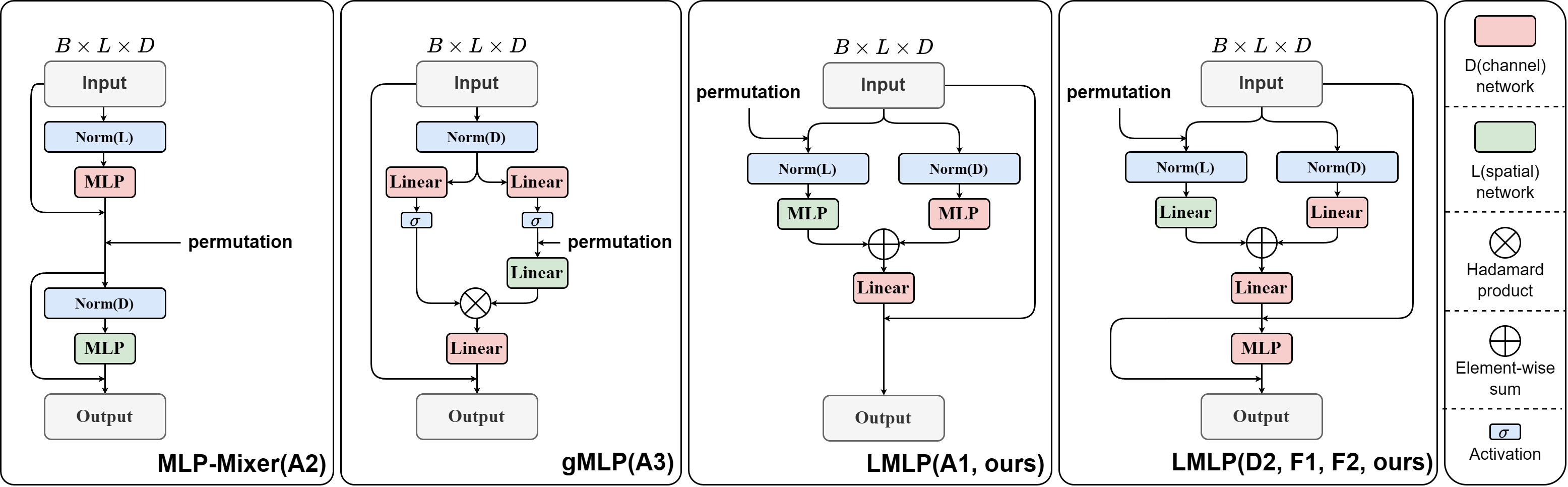}
  \caption{Block design comparison with MLP-Mixer and gMLP. For all designs, see Appendix A.2.}
  \label{fig:design}
\end{figure}

L-MLP architecture is subject to design choices that can impact the performance significantly. 
Similar to the case of the attention mechanism, where the design details are discovered through empirical explorations, we also run a series of experiments on several model designs to find the best L-MLP architecture. These models are trained for 100K steps ($1/10$ of the main experiments) due to limited computational resources. 
Our model design starts from a simple parallel setup (A1) (see Figure \ref{fig:design}), where $\textit{FNN}_L$ and $\textit{FNN}_R$ are 1-layer MLPs, merging mechanism is Add + Linear, and $\textit{FNN}_Z$ is not used. All models in the same group are tuned to be comparable in terms of the number of parameters. These experiments can also serve as ablative experiments to demonstrate our optimal design.

\begin{figure}[h!]
  \centering
  \includegraphics[width=\textwidth]{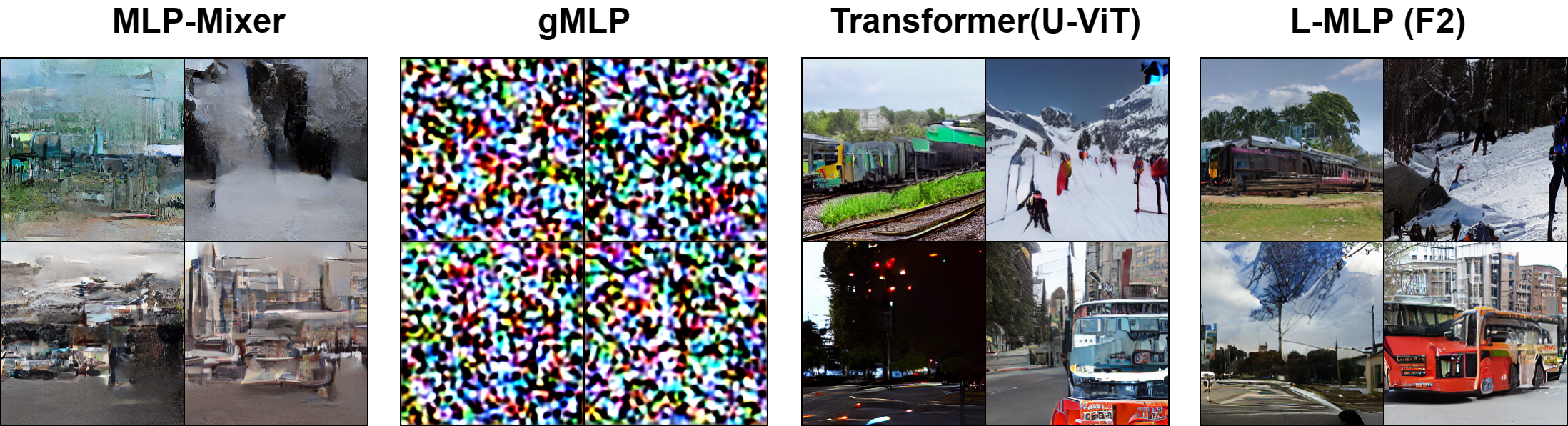}
  \caption{Generated images from text prompts (From left to right, top to bottom: `A green train is coming down the tracks.', `A group of skiers are preparing to ski down a mountain.', `A road with traffic lights, street lights, and cars.', `A bus driving in a city area with traffic signs.') at 100K steps for different models. We compare our L-MLP (F2) with two MLP variants and Transformers (U-ViT). }
  \label{fig:genimages}
\end{figure}

In Table \ref{tab1} (a), we show that the basic L-MLP outperform MLP-Mixer and gMLP significantly, while the other two MLP-based models fail to generate meaningful images. In Table \ref{tab1}  (b) to (f), we finalized our best model design to be F2, where $\textit{FNN}_L$ and $\textit{FNN}_R$ are Linear projections, merging mechanism is Add + Linear, $\textit{FNN}_Z$ is a one-layer MLP, and inter blocks skip connections are arranged in a U-shaped manner similar to that of a U-Net. In addition, in Table \ref{tab1}  (b), we show that the projection after merging is crucial for the performance from the drop of performance from A1 to B3. We also tested several GLU setups (B2, E1, E2), since GLUs have recently been adopted in similar tasks~\cite{crowson2024scalable}. However, these variants fail to achieve comparable performance to L-MLP-based designs. We visualize the generated samples by 4 representative models in Figure~\ref{fig:genimages}. It shows that the L-MLP is the only MLP-based model that can generate comparable images to the Transformer-based counterpart.

\begin{table}[h]
\footnotesize
\centering
\caption{Architectural design choices based on generation quality: different design choices are used to train the model under the same configuration through 100K training steps. Model parameters are controlled to be comparable in each group by adjusting model hyperparameters. F2-deep is the final model we used to perform our comparative analyses.}

\begin{tabular}{cc}
\begin{minipage}[t]{.45\linewidth}
\vspace{0pt}
\centering
\subcaption{MLP variants}
\begin{threeparttable}
\begin{tabular}{l|lccc}
\toprule
\multicolumn{2}{l}{\textbf{Configuration}} & \textbf{FID$\downarrow$} & \textbf{\#Params} \\
\midrule
A1 & L-MLP (ours) & \textbf{36.17} & 23.5M \\
A2 & MLP-Mixer~\cite{Tolstikhin2021MLPMixerAA} &  >100\tnote{*} & 23.5M\\
A3 & gMLP~\cite{Liu2021PayAT} &  >100 & 23.5M \\
\bottomrule
\end{tabular}
\begin{tablenotes}
    \item[*] $>100$ indicates the generated quality is too poor that subjects are visually meaningless.
\end{tablenotes}
\end{threeparttable}
\end{minipage} &
\begin{minipage}[t]{.45\linewidth}
\vspace{0pt}
\centering
\subcaption{Merging Mechanism}
\begin{threeparttable}
\begin{tabular}{l|lccc}
\toprule
\multicolumn{2}{l}{\textbf{Configuration}} & \textbf{FID$\downarrow$} & \textbf{\#Params }  \\
\midrule
A1 & Add + Proj& \textbf{36.17} & 23.5M\\
\midrule
B1 & Prod\tnote{*} + Proj&  37.21 & 23.5M \\
B2 & GLUs + Proj&  38.69 & 23.5M \\
B3 & Add + None&  >100 & 23.5M \\

\bottomrule
\end{tabular}
\begin{tablenotes}
    \item[*] Element-wise product.
\end{tablenotes}
\vspace{-8mm}

\end{threeparttable}

\end{minipage} \\
\vspace{-4mm}
\begin{minipage}[t]{.45\linewidth}
\centering
\subcaption{Second Stage Variants}
\begin{threeparttable}
\begin{tabular}{l|lccc}
\toprule
\multicolumn{2}{l}{\textbf{Configuration}} & \textbf{FID$\downarrow$} & \textbf{\#Params} \\
\midrule
A1 & None & 24.85 & 45M\\ 
\midrule
C1 & MLP &  \textbf{20.07} & 45M  \\
\bottomrule
\end{tabular}
\end{threeparttable}
\end{minipage} &

\begin{minipage}[t]{.45\linewidth}
\centering
\subcaption{First Stage Variants}
\begin{tabular}{l|lccc@{}}
\toprule
\multicolumn{2}{l}{\textbf{Configuration}} & \textbf{FID$\downarrow$} & \textbf{\#Params}  \\
\midrule
C1 & MLP & 20.07 & 45M\\
\midrule
D1 & Linear + Prod & 18.74 & 45M \\
D2 & Linear + Sum & \textbf{18.01} & 45M \\

\bottomrule
\end{tabular}
\end{minipage} \\

\begin{minipage}[t]{.45\linewidth}
\centering
\subcaption{First Stage Activation}
\begin{threeparttable}
\begin{tabular}{l|lccc}
\toprule
\multicolumn{2}{l}{\textbf{Configuration}} & \textbf{FID$\downarrow$} & \textbf{\#Params}  \\
\midrule
D2 & None + Sum & \textbf{18.01} & 45M \\
\midrule
E1 & GELU\tnote{*} + Prod & 20.24 & 45M  \\
E2 & GELU + Sum & 19.48 &  45M \\
\bottomrule
\end{tabular}
\begin{tablenotes}
    \item[*] GELU is only used on the left branch (permuted input) after the linear layer.
\end{tablenotes}
\end{threeparttable}

\end{minipage} &

\begin{minipage}[t]{.45\linewidth}
\centering
\subcaption{U-Net Skip Connections}
\begin{threeparttable}
\begin{tabular}{l|lccc}
\toprule
\multicolumn{2}{l}{\textbf{Configuration}} & \textbf{FID$\downarrow$} & \textbf{\#Params}  \\
\midrule
D2 & None & 18.01 & 45M \\
\midrule
F1 & First Stage & 16.63 & 45M\\
F2\tnote{\textdagger} & Second Stage & 16.13 & 45M\\
F2-Deep\tnote{*} & Second Stage & \textbf{13.66} & 47M\\

\bottomrule
\end{tabular}
\begin{tablenotes}
    \item[\textdagger] MLP scale $=5.2$ 
    \item[*] 4 more L-MLP layers than F2, MLP scale $=4$      
\end{tablenotes}
\end{threeparttable}
\end{minipage}
\end{tabular}
\label{tab1}
\end{table}

\subsection{Main Comparative Results}
\begin{figure}[h] 
    \centering 
    \includegraphics[width=\linewidth]{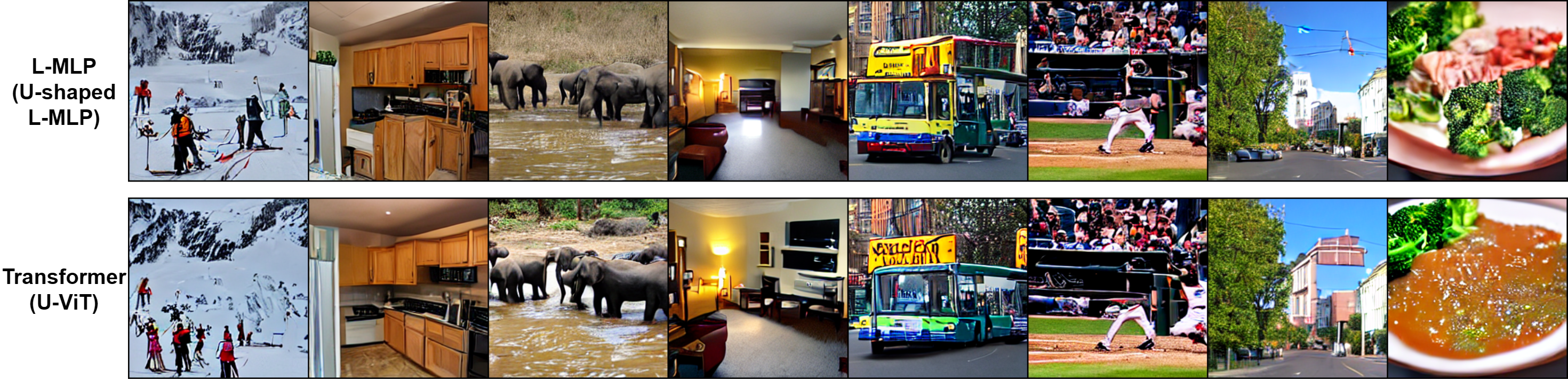} 
    \caption{Generated samples from validation prompts. Both models share similar visual features. See Appendix A.4 for more comparisons.}
    \label{fig:samples}
\end{figure}

We compare the quality of generated images by our architecture with prominent models with comparable sizes. Our results are presented in Table~\ref{tab:models_coco}. We observe that our model is able to reach a FID score of 8.62, which is the best-performing MLP-based backbones and comparable with attention-based backbones. It is crucial to note that attention-based models have improved since the introduction of the attention mechanism and we envision that our L-MLP-based model can be improved further in time. Despite the performance gap, our results suggest for the first time that the attention mechanism or convolution-based block designs are not essential to solving complex generative tasks.  For visual inspection, we have also included a few examples of generated images using validation prompts by the models in Figure \ref{fig:samples} (for more samples, please refer to Appendix A.4).  We see that when using the same random seed, the UL-MLP generates similar images to those generated by U-ViT. This is an interesting result since the fully MLP-based model shows similar learning behavior across $50$ sample steps, while not using the attention mechanism.

\begin{figure}[h] 
    \centering 
    \includegraphics[width=\textwidth]{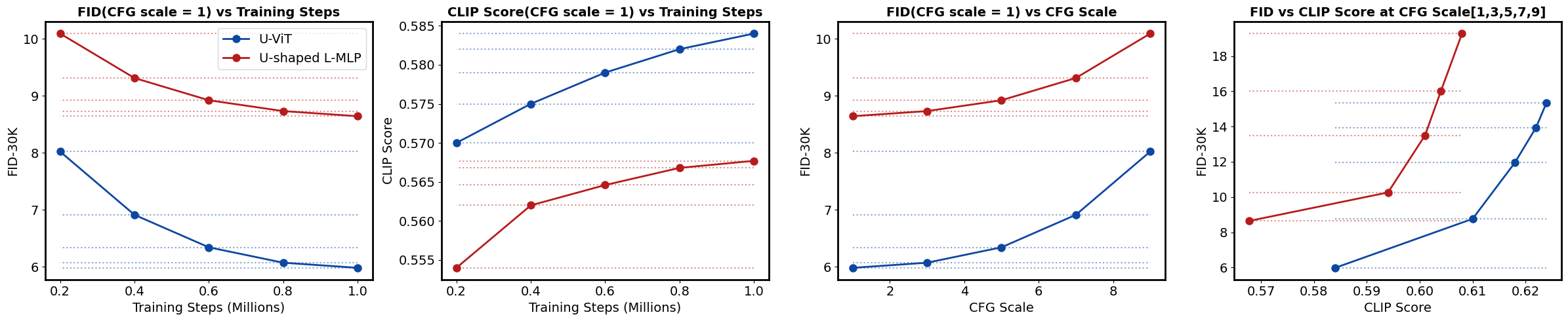} 
    \caption{FID and CLIP scores comparison with U-ViT-S/2. The UL-MLP shows similar training behavior as the Transformer-based counterpart while converging slightly faster.}
    \label{fig:curves}
\end{figure}

We have also compared UL-MLP to U-ViT in Figure~\ref{fig:curves} based on the FID and CLIP scores. We observe that L-MLP is exhibiting similar training behaviors as the U-ViT: its curves follow similar patterns as the U-ViT. Admittedly, there's still a gap between L-MLP and Transformer, but this training behavior demonstrates similar scalability and stable training to Transformer on generative tasks, which none of the existing MLP-based architecture has shown the capability of. Both Figure~\ref{fig:samples} and Figure~\ref{fig:curves} provide evidence that L-MLP provides similar functionalities as the attention mechanism.

\begin{table}[ht]
\footnotesize
\centering
\caption{Generative models trained on MS-COCO dataset with their FID scores, types, architectural details, and parameter counts.}
\label{tab:models_coco}
\begin{threeparttable}
\begin{tabular}{lcccc}
\toprule
\textbf{Model}         & \textbf{FID}$\downarrow$ & \textbf{Type}           & \textbf{Architecture} & \textbf{\#Params}\tnote{a}  \\ 
\midrule
AttnGAN~\cite{Rombach2021HighResolutionIS}           & 35.49        & GAN                     & CNN + Attention                  & 230M              \\ 
DM-GAN~\cite{Zhu2019DMGANDM}            & 32.64        & GAN                     & CNN + Memory Network                   & 46M               \\ 
VQ-Diffusion~\cite{Gu2021VectorQD}     & 19.75        & Diffusion      & CNN + Transformer                   & 370M              \\ 
XMC-GAN~\cite{Zhang2021CrossModalCL}         & 9.33         & GAN                     & CNN + Attention                                & 166M              \\ 
LAFITE~\cite{zhou2021lafite}          & 8.12         & GAN                     & CNN + Transformer                  & 75M + 151M    \\ 
LDM~\cite{Rombach2021HighResolutionIS}                  & 7.32         & Latent diffusion        & CNN + Cross-attention                   & 53M + 207M      \\ 
U-ViT-S/2~\cite{Bao2022AllAW}              & 5.95         & Latent diffusion        & CNN + Transformer           & 45M + 207M     \\ 
IF-U-ViT~\cite{hu2024intermediate}              & 5.68         & Latent diffusion        & CNN + Transformer           & 48M + 207M    \\ 
U-ViT-S/2 (Deep) ~\cite{Bao2022AllAW}         &\textbf{ 5.48 }        & Latent diffusion        & CNN + Transformer           & 58M + 207M    \\ 
\midrule
\multicolumn{5}{l}{\textbf{MLP-based}\tnote{b}} \\
\midrule
gMLP~\cite{Liu2021PayAT}\tnote{*}       & >100         & Latent diffusion        & CNN + MLP           & 45M + 207M     \\ 
MLP-Mixer~\cite{Tolstikhin2021MLPMixerAA}\tnote{*}      & >100         & Latent diffusion        & CNN + MLP           & 45M + 207M     \\ 
    UL-MLP (F2-Deep, ours)       & \textbf{8.62}        & Latent diffusion        & CNN + MLP           & 47M + 207M      \\ 
\bottomrule
\end{tabular}
\begin{tablenotes}
    \item[a] $+$ indicated the frozen \#Params of pretrained encoders and decoders.
    \item[b] CNN is used only for pre- and post-processing, which is not included in the main block design.
    \item[*] Our implementation on the diffusion task. 
\end{tablenotes}
\end{threeparttable}
\end{table}

\section{Analytic Insights}

\subsection{Efficiency comparison with transformers}
Transformers follow two-stage processing: (i) embedding sequential information into the feature space using scaled dot-product attention. (ii) using the feature space MLP to process the information. L-MLP also follows a similar two-stage design but it relaxes the number of dot-product operations by three in the first stage, improving the inference speed. Table~\ref{tab:block_eff} shows the complexity comparison between our architecture and an attention block.  L-MLP has a higher GMACs-size ratio than a transformer block which indicates that L-MLP is more hardware-efficient at scale. This property is highly appealing because currently hardware resource is a major obstacle for scaling transformers.

\begin{table}[h!]
    \centering
    \footnotesize
    \caption{Efficiency Comparison between Transformer block and L-MLP block}
    \begin{threeparttable}
    \begin{tabular}{lcccc}
    \toprule
    \textbf{Model} & \textbf{Complexity}\tnote{a} & \textbf{\#Params(P)} & \textbf{GMACs(M)} & \textbf{M/P Ratio}  \\
    \midrule
    Transformer block & $(4 + 2s)LD^2 + 2 L^2D$ & $(4 + 2s)D^2$ & - & -\\
    L-MLP block & $(2 + 2s)LD^2 + L^2D$ & $(2+2s)D^2 + L^2$ & - & -\\
    \midrule
    Transformer block (s=4)& $\approx1.165$B & $\approx 3.42$M  & $1.05$G & $\approx307.02$ \\
    L-MLP block (s=5.2)\tnote{b} & $\approx1.143$B & $3.42$M & $1.16$G  & $ 339.18$ \\
    L-MLP block (s=4)\tnote{c} & $\approx0.933$B & $2.74$M  & $0.935$G & $ 341.24$\\
    \bottomrule
    \end{tabular}
    \begin{tablenotes}
    \item[a] A single forward pass. Lower-order terms such as bias and normalization are omitted.
    \item[b] Used in L-MLP (F2). s is chosen to match the \#Params of the default Transformer block.
    \item[c] Used in L-MLP (F2-deep).
    \end{tablenotes}
    \end{threeparttable}
    \label{tab:block_eff}
\end{table}

We further empirically look at the efficiency of the full-size diffusion backbone models in Table~\ref{tab:model_eff}. We observe that despite being larger than U-ViT, the L-MLP is faster in terms of GMACs, training speed, and inference speed, due to the faster operations of MLP-based block design. We think this is only a start and MLP-based architecture has the potential to become more efficient and scalable.

\begin{table}[h!]
    \centering
    \footnotesize
    \caption{Efficiency Comparison between U-ViT and UL-MLP}
    \begin{threeparttable}
    \begin{tabular}{lcccc}
    \toprule
    \textbf{Model} & \textbf{\#Params (P)} & \textbf{GMACs} & \textbf{Training Speed}\tnote{a} & \textbf{Inference Speed}\tnote{b} \\
    \midrule
    U-ViT& $45$M & $14.78$G & $1.81$ steps/s & $7.0$ imgs/s \\
    UL-MLP (F2-deep) & $47$M & $15.95$G & $2.15$ steps/s & $9.8$ imgs/s \\
    \bottomrule
    \end{tabular}
    \begin{tablenotes}
    \item[a] A EMA training step with a batch size of 256 on an RTX 4090 GPU. 
     \item[b] DPM-Solver CFG sampling with 50 steps, batch size of 20. Averaged with file saving. 
    \end{tablenotes}
    \end{threeparttable}
    \label{tab:model_eff}
\end{table}

\subsection{Network analysis}
\begin{figure}[h!] 
    \centering 
    \includegraphics[width=\textwidth]{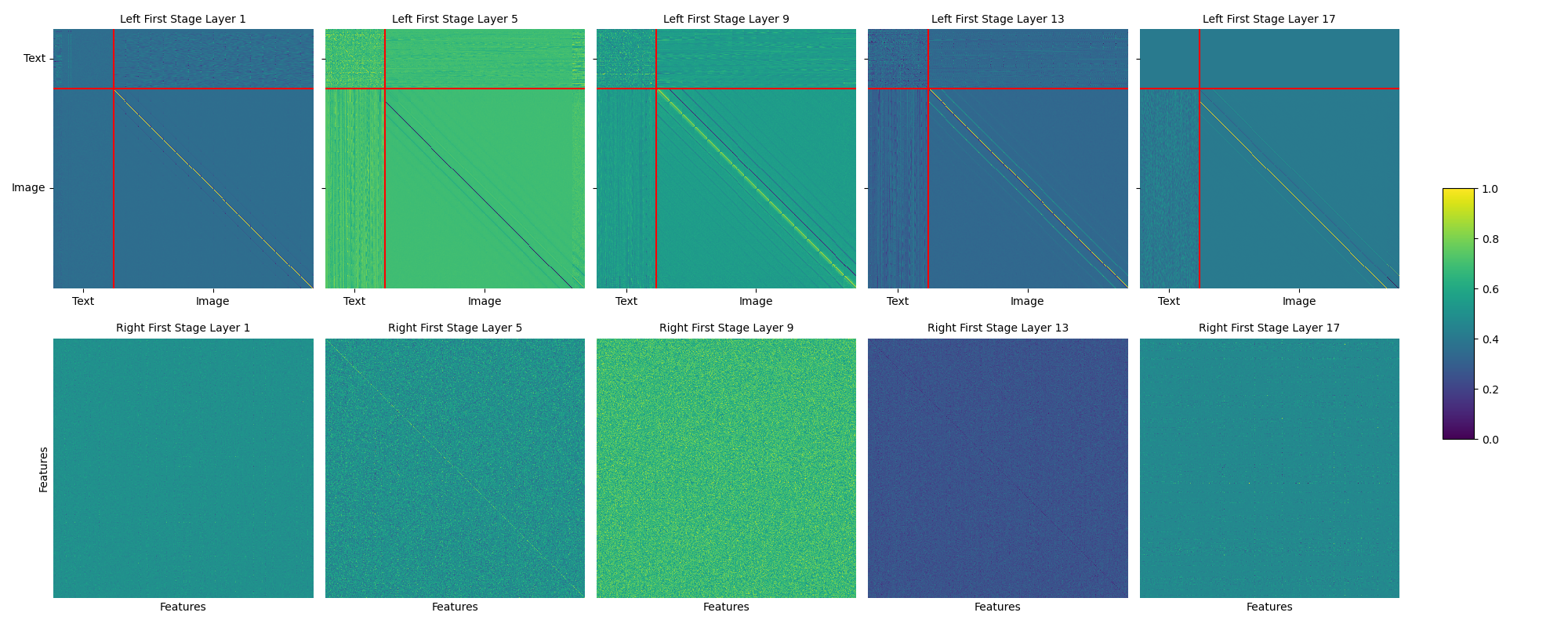} 
    \caption{First stage linear layers weight visualizations. We select Layers 1, 5, 9, 13, and 17 to visualize the weights of the left and right hemisphere linear layers. The left and right linear layers are exhibiting functional lateralization. (Best viewed when zoomed-in)}
    \label{fig:weights}
\end{figure}

We further investigate the weight patterns of the first-stage linear layers in Figure~\ref{fig:weights} to gain insights. All weights are linearly normalized to the range of $[0, 1]$. The x-axis denotes the input dimension, and the y-axis denotes the output. We observe several behaviors that are similar to the attention maps of transformer-based backbones. A previous study shows that U-ViT tends to have high-ranked attention close to the middle layers~\cite{hu2024intermediate}. The L-MLP first stage also exhibits similar behaviors by having higher weights in the middle layer, while having less prominent weights close to the early and late layers.  The first row corresponds to the left-hemisphere linear layers for the permuted input.  We see that the four separated regions (text-to-text, image-to-text, text-to-image, and image-to-image), separated by the red lines,  have distinct behaviors. The text-to-image, image-to-text, and image-to-image regions show repeated patterns 16 times, which is the number of rows of the image patches. These patterns serve a similar role as the positional embedding used in the transformers. The diagonal lines in the image-to-image regions reflect image patch-wise interactions. We see that the patches that are spatially close to each other (repeated diagonal lines due to neighbors in the column direction after row break) have more distinct interactions. This behavior indicates the left-hemisphere linear layers capture the spatial property of images. The second row corresponds to the right-hemisphere linear layers. We see that the repeated patterns and diagonals are less prominent, reflecting the focus of this region is less spatial, but more on the details of the features. These preliminary observations indicate that L-MLP shows behaviors similar to the nervous system, where functional asymmetry happens during learning. In conclusion, the L-MLP's first stage achieves functional lateralization through a simple model design which is achieved by processing different dimensions separately while maintaining the consistency of other dimensions.

\section{Limitations and Future Work}
As observed in the experiments,  L-MLP still has an expressive gap compared to the transformer-based models. This gap might be due to the high-order product involved in the Transformer which enriches its expressiveness. Although L-MLP is aligned with the human brain's functionality, where inputs are limited to fixed-scope sensory systems, the current design scales quadratically to a fixed sequence length, limiting its usage in natural language processing. Additional designs that enable memory-efficient inference, or dynamic latent space are needed to process longer and dynamic sequences with subquadratic scaling. 
For future research directions, convolutional networks and recurrent networks can be introduced in our architecture, replacing the linear layers to introduce additional inductive bias more suitable for image and text data. Although higher-order interactions such as product-based gates such as gMLP and GLU exhibit weaker performance in our design experiments when training steps are limited, further investigation on the full-training behavior is needed. These directions have the potential to help resolve the remaining expressive gap between the Transformers and L-MLP.

\section{Conclusion}
We introduced the L-MLP as a compelling alternative to transformer-based architectures, inspired by the functional asymmetry of the human brain. By using a unique architecture that permutes and processes data dimensions in parallel, the L-MLP effectively closes the gap between MLP-based models and the more complex transformer-based models in the context of challenging text-to-image diffusion tasks. Our experiments on the MS-COCO dataset reveal that the L-MLP not only achieves competitive performance with lower computational costs but also aligns well with biological insights into brain function, suggesting a promising direction for developing efficient and brain-inspired computational models. The network's ability to mimic lateralization observed in human neural processes underscores its potential to enhance machine learning models' efficiency and effectiveness without relying on the computationally expensive mechanisms inherent to transformers.

{
\small
\bibliography{reference}
\bibliographystyle{plain}
}

\appendix
\clearpage
\section{Appendix / supplemental material}
\subsection{Main model training settings}
We provide the training settings for our main experiments. These settings are not fine-tuned for our L-MLP model and are originally used by U-ViT.
\begin{table}[h]
\centering
\caption{Model Configuration}
\begin{tabular}{lll}
\hline
\textbf{Component} & \textbf{Parameter} & \textbf{Value} \\ \hline
\textbf{Autoencoder} & Scale Factor & 0.23010 \\ \hline
\textbf{Training} & Number of Steps & 2,000,000 \\
 & Batch Size & 128 \\
 & Accumulation Steps & 2 \\ \hline
\textbf{Device} & GPU & 1 $\times$ \text{NVIDIA RTX 4090} \\ \hline
\textbf{Optimizer} & Type & AdamW \\
 & Learning Rate & 0.0002 \\
 & Weight Decay & 0.03 \\
 & Betas & (0.9, 0.9) \\ \hline
\textbf{LR Scheduler} & Type & Linear \\
 & Warmup Steps & 5,000 \\ \hline
\textbf{Neural Network} & Variation & F2-deep \\
 & Image Size & 32 \\
 & Input Channels & 4 \\
 & Patch Size & 2 \\
 & Embedding Dimension & 512 \\
 & Depth & 16 \\
 & MLP Ratio & 4 \\
 & CLIP Dimension & 768 \\
 & Solver & DPM-Solver\\
 & Number of CLIP Tokens & 77 \\ \hline
\textbf{Sampling} & Sample Steps & 50 \\
 & CFG Scale & 1.0 \\ \hline
\end{tabular}
\end{table}

\subsection{Pseudo-code for L-MLP}
We provide the pseudo-code for the L-MLP block. Here all linear layers are shape-preserving (square metrics). 
\begin{algorithm}
\caption{Forward pass of the proposed L-MLP}
\begin{algorithmic}[1]
    \REQUIRE $x$: input tensor of shape (B, L, D)
    \STATE $x_{\text{perm}} \leftarrow \text{x.rearrange}(\text{`B L D} \rightarrow \text{B D L'})$
    \STATE $x \leftarrow \text{norm}(x, \text{dim} = \text{D})$
    \STATE $x_{\text{perm}} \leftarrow \text{norm}(x_{\text{perm}}, \text{dim} = \text{L})$
    \STATE $l \leftarrow \text{Linear}(x_{\text{perm}})$
    \STATE $l \leftarrow \text{l.rearrange}(l, \text{`B D L} \rightarrow \text{B L D'})$
    \STATE $r \leftarrow \text{Linear}(x)$
    \STATE $x \leftarrow x + \text{Linear}(l + r)$
    \STATE $x \leftarrow \text{norm}(x,  \text{dim} = \text{D})$
    \STATE $x \leftarrow x + \text{MLP}(X)$
    \RETURN $x$
\end{algorithmic}
\end{algorithm}

\subsection{Design choices}
\begin{figure}[htbp]
    \centering
    \includegraphics[width=\linewidth]{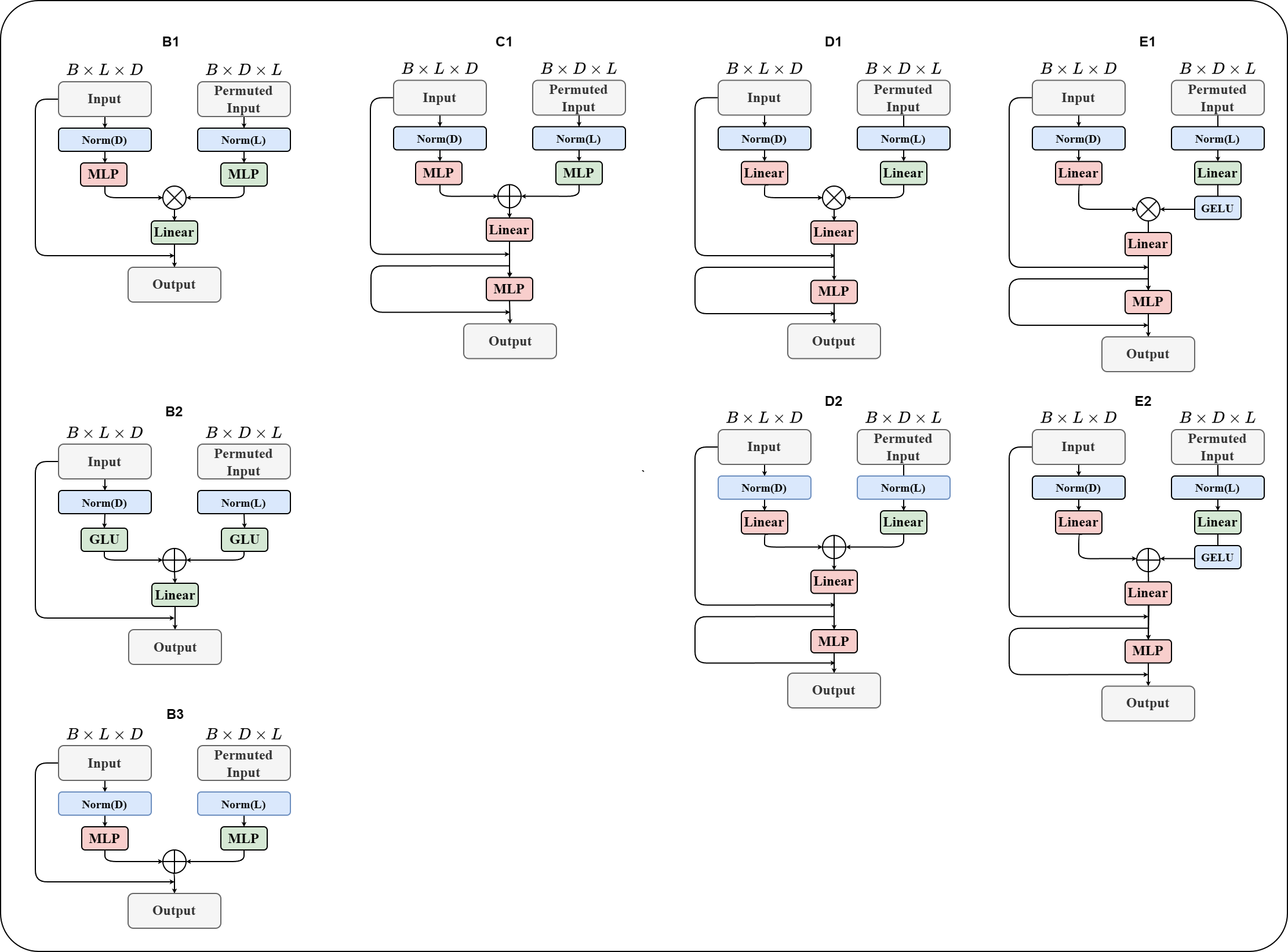}
    \caption{Other design choices.}
    \label{fig:design_choices}
\end{figure}

We provide all design choices tested in our design ablations in Figure~\ref{fig:design_choices}, this is a supplement for Figure~\ref{fig:design}, and covers all scenarios in Table~\ref{tab1}. 

\subsection{More comparative samples}
We provide more comparisons between the generated images by UL-MLP and U-ViT model.
\begin{figure}[htbp]
    \centering
    \begin{subfigure}{0.5\textwidth}
        \centering
        \includegraphics[width=0.98\linewidth]{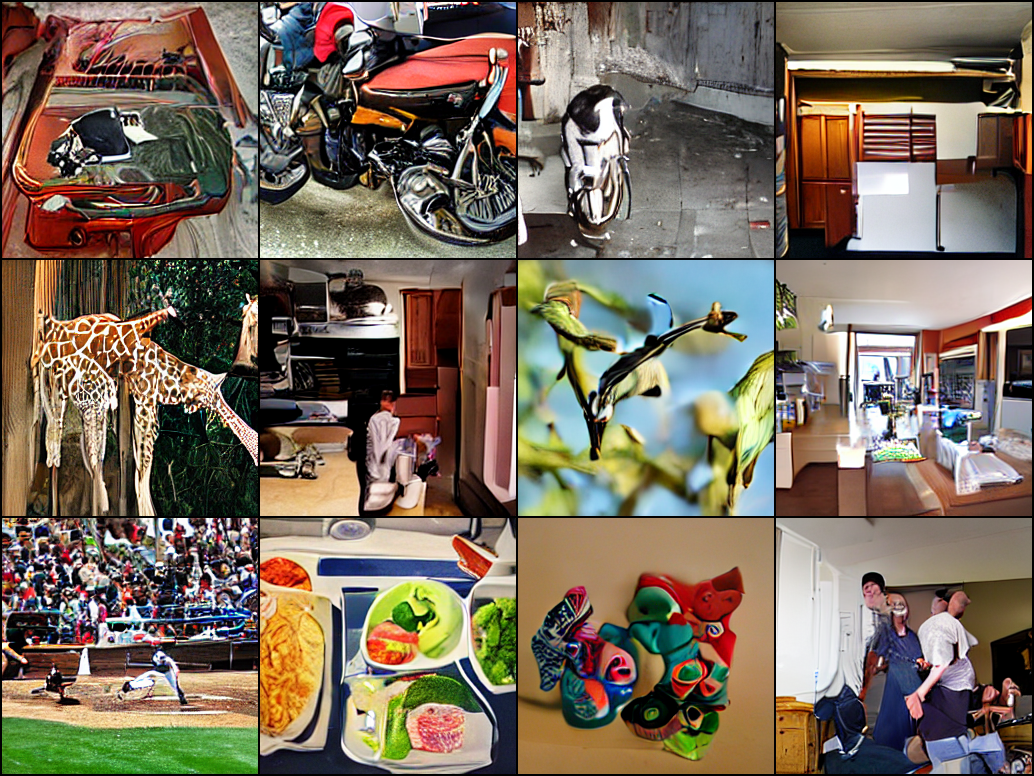}
        \caption{L-MLP generation for CFG scale 1}
    \end{subfigure}%
    \begin{subfigure}{0.5\textwidth}
        \centering
        \includegraphics[width=0.98\linewidth]{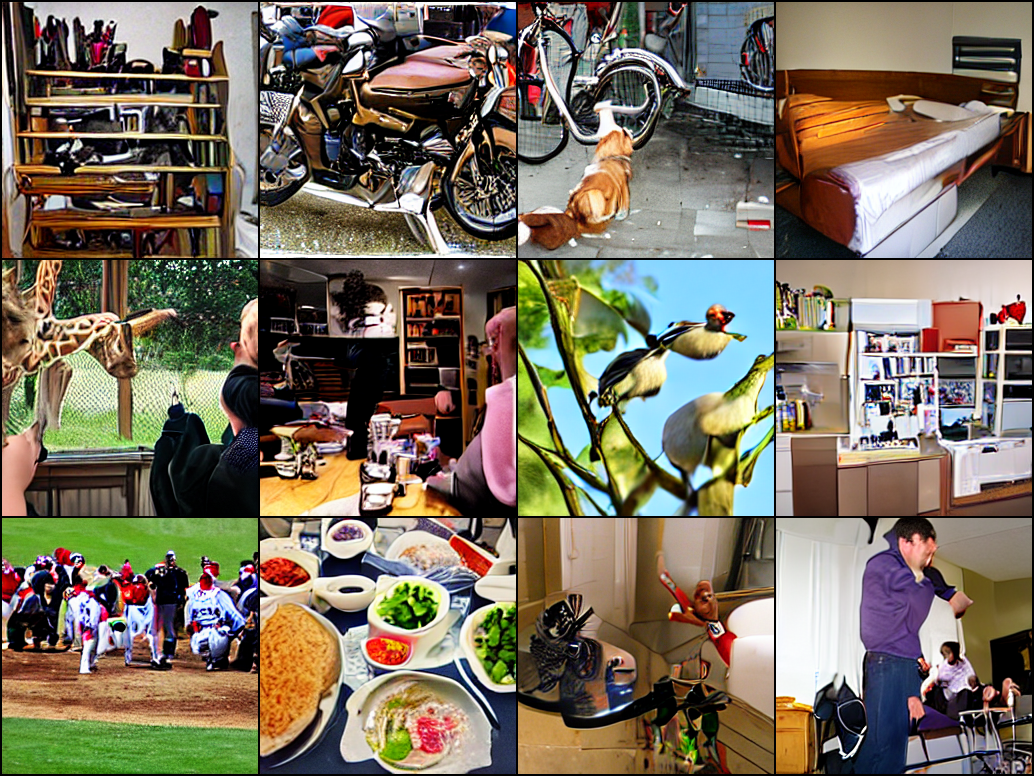}
        \caption{U-ViT generation for CFG scale 1}
    \end{subfigure}

    \begin{subfigure}{0.5\textwidth}
        \centering
        \includegraphics[width=0.98\linewidth]{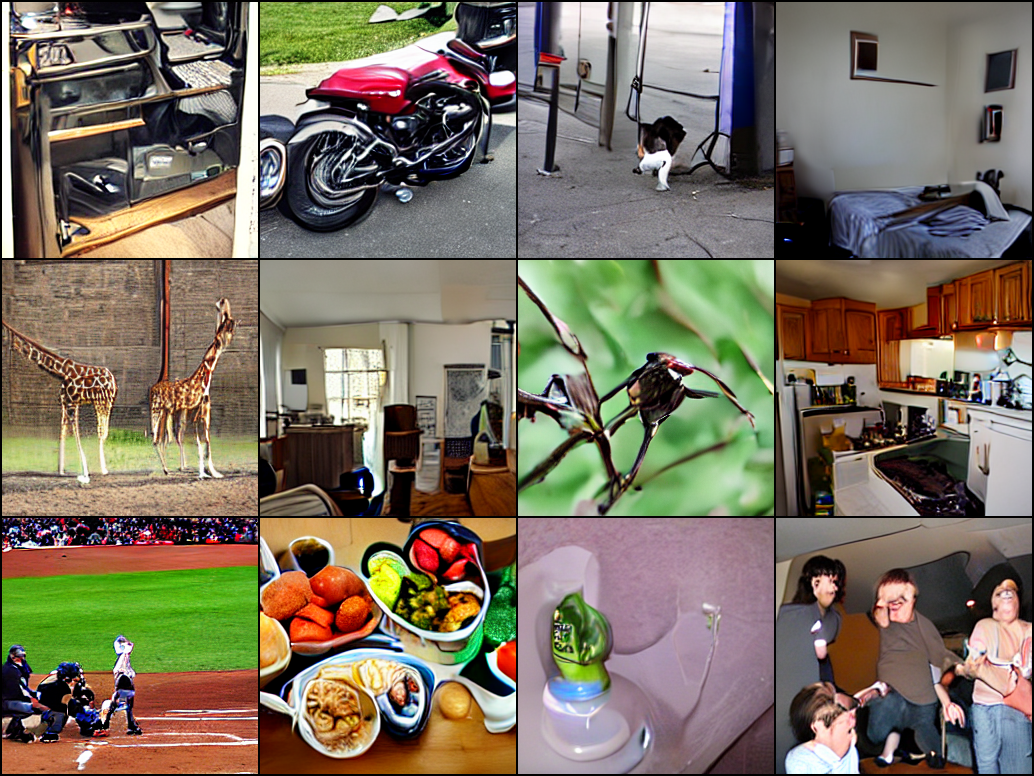}
        \caption{L-MLP generation for CFG scale 3}
    \end{subfigure}%
    \begin{subfigure}{0.5\textwidth}
        \centering
        \includegraphics[width=0.98\linewidth]{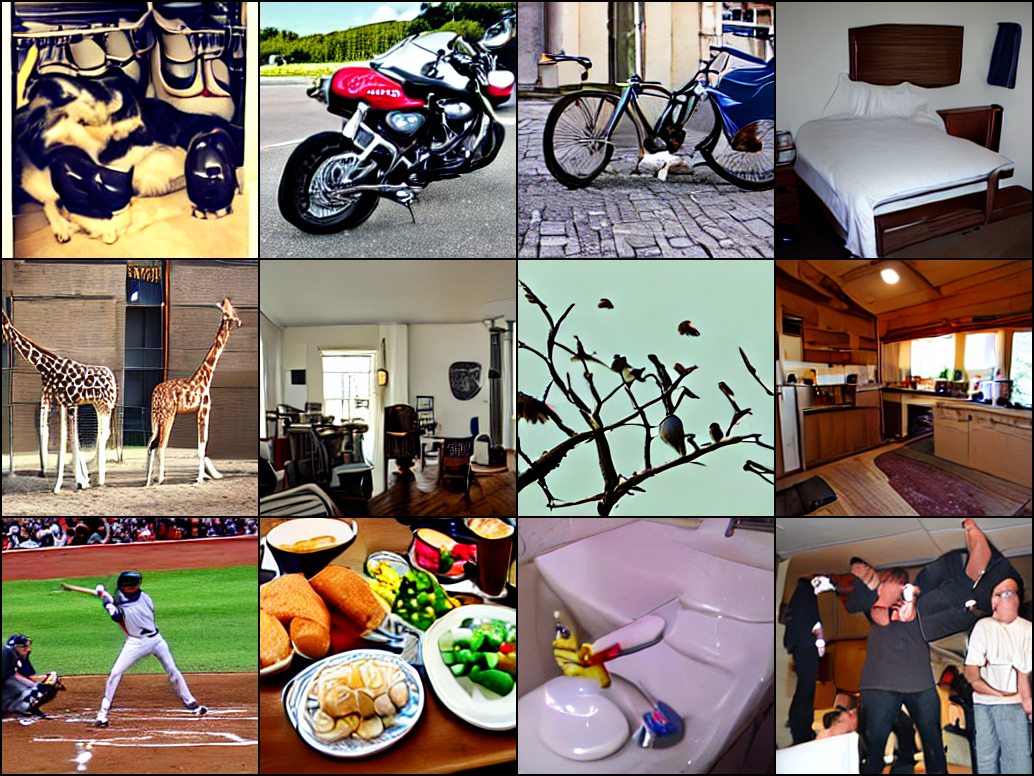}
        \caption{U-ViT generation for CFG scale 3}
    \end{subfigure}

    \begin{subfigure}{0.5\textwidth}
        \centering
        \includegraphics[width=0.98\linewidth]{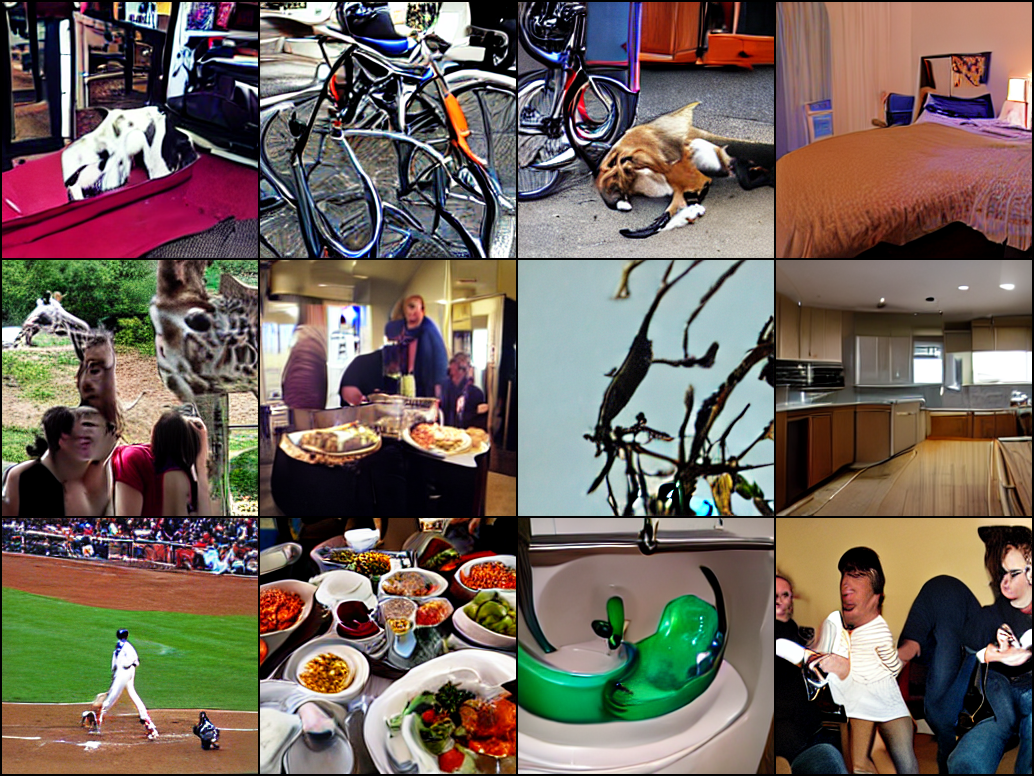}
        \caption{L-MLP generation for CFG scale 5}
    \end{subfigure}%
    \begin{subfigure}{0.5\textwidth}
        \centering
        \includegraphics[width=0.98\linewidth]{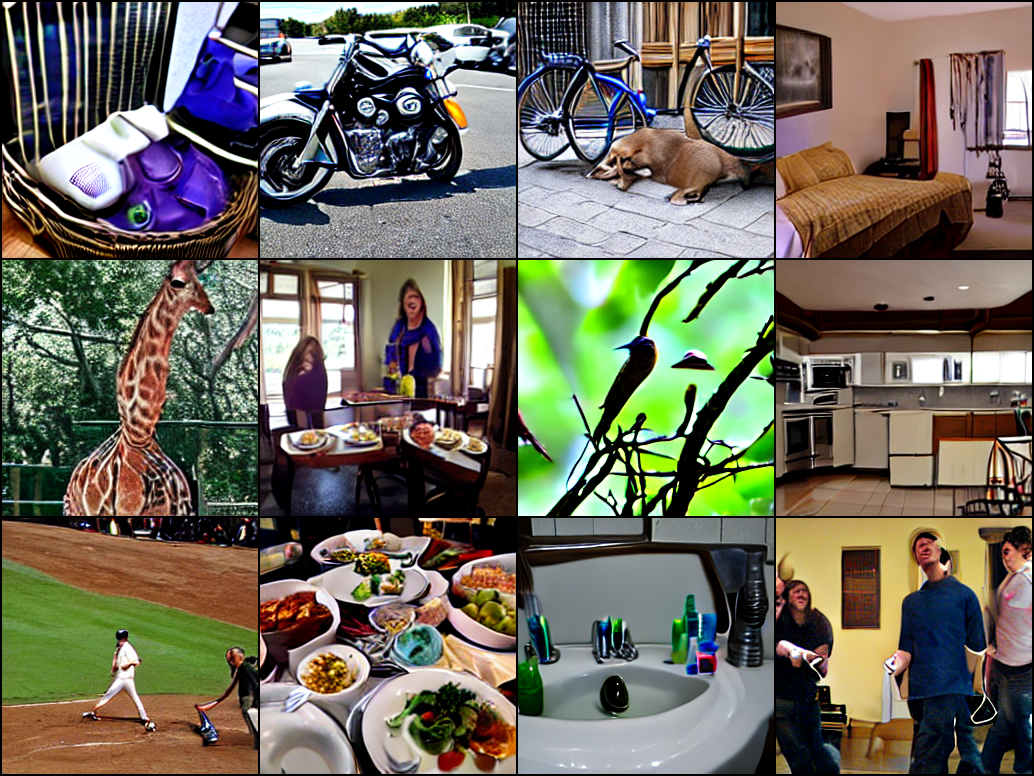}
        \caption{U-ViT generation for CFG scale 5}
    \end{subfigure}

    \begin{subfigure}{0.5\textwidth}
        \centering
        \includegraphics[width=0.98\linewidth]{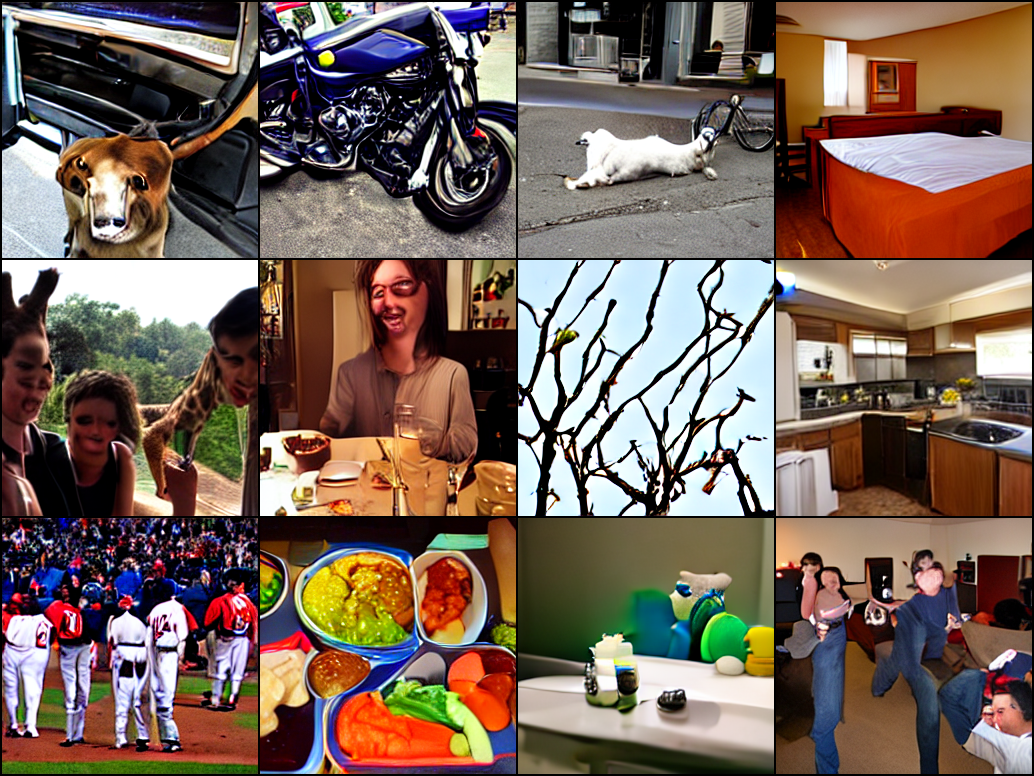}
        \caption{L-MLP generation for CFG scale 7}
    \end{subfigure}%
    \begin{subfigure}{0.5\textwidth}
        \centering
        \includegraphics[width=0.98\linewidth]{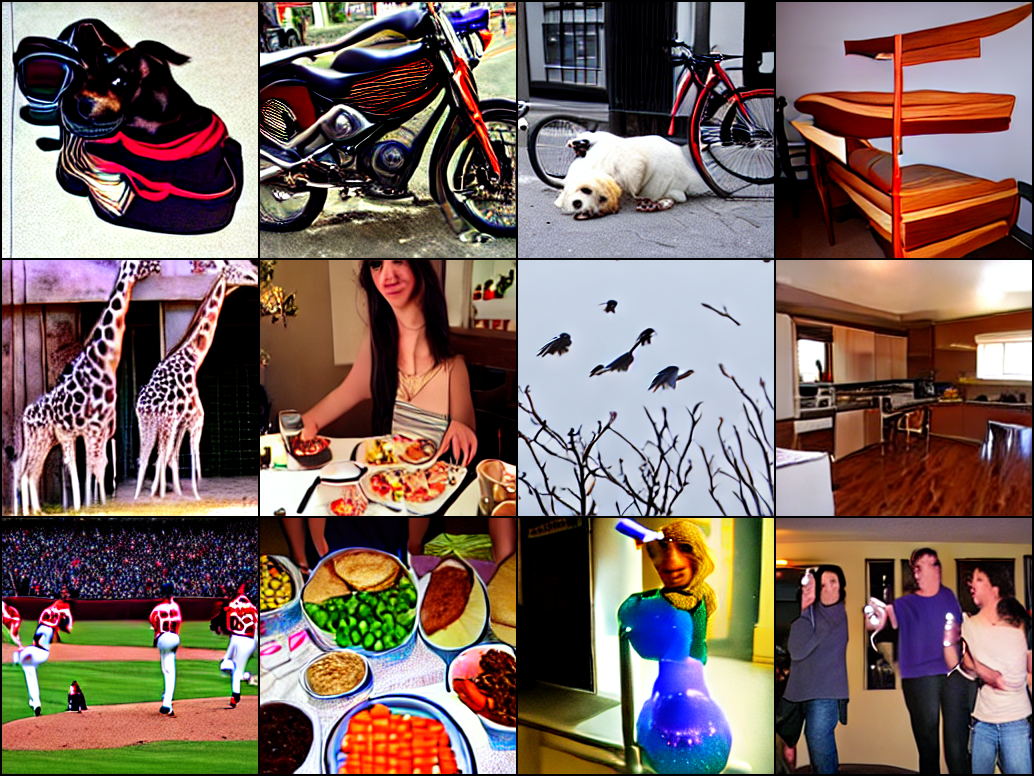}
        \caption{U-ViT generation for CFG scale 7}
    \end{subfigure}
    
    \caption{Comparison between L-MLP generation and UViT generation across different CFG scales.}
\end{figure}
\end{document}